\documentclass[twoside]{article}
\usepackage{ArXiv_sty}
\usepackage{bm}
\usepackage{amsmath}
\usepackage{amssymb}
\usepackage{amsthm}
\usepackage{algorithm}
\usepackage{algpseudocode}
\usepackage{graphicx}
\usepackage{epstopdf}
\usepackage{subfig}
\usepackage{placeins}
  \usepackage{color}
\usepackage{wrapfig}

\newcommand{\RR}{\mathbb{R}}%
\newcommand{\DD}{\mathcal{D}}

\newcommand{\UU}{\mathcal{U}}
\DeclareMathOperator*{\argmax}{argmax}
\DeclareMathOperator*{\argmin}{argmin}

\usepackage{pifont}
\newcommand{\cmark}{\ding{51}}%
\newcommand{\xmark}{\ding{55}}%

\newtheorem{definition}{Definition}

\newtheorem{remark}{Remark}

\usepackage[round]{natbib}  
\begin{document}

\twocolumn[

\aistatstitle{Advancing Bayesian Optimization: The Mixed-Global-Local (MGL) Kernel and Length-Scale Cool Down}

\aistatsauthor{Kim Peter Wabersich \And Marc Toussaint}

\aistatsaddress{Machine Learning and Robotics Lab,\\University of Stuttgart \And Machine Learning and Robotics Lab,\\University of Stuttgart} ]

\begin{abstract}
Bayesian Optimization (BO) has become a core method for solving
expensive black-box optimization problems. While much research
focussed on the choice of the acquisition function, we focus on online
length-scale adaption and the choice of kernel function. Instead of
choosing hyperparameters in view of maximum likelihood on past data,
we propose to use the acquisition function to decide on hyperparameter
adaptation more robustly and in view of the future optimization
progress. Further, we propose a particular kernel function that
includes non-stationarity and local anisotropy and thereby implicitly
integrates the efficiency of local convex optimization with global
Bayesian optimization. Comparisons to state-of-the art BO methods
underline the efficiency of these mechanisms on global optimization
benchmarks.
\end{abstract}

\section{INTRODUCTION}\label{Sec:Introduction}

Bayesian Optimzation (BO) became an almost ubiquitous tool for general
black-box optimization with high function evaluation cost. The key
idea is to make use of the gathered data by computing the Bayesian
posterior over the objective function. A BO algorithm is in principle
characterized by two choices: 1) What is the prior over the objective
function? 2) Given a posterior, what is the decision theoretic
criterion, the so-called acquisition function, to choose the next query
point?

Previous research has extensively focussed on the second question.
For instance, recent work goes beyond
Expected Improvement \citep{jones2001taxonomy} and UCB
\citep{srinivas2012information} by proposing the
entropy of the optimum location \citep{hernandez2014predictive} or
an infinite metric based criterion
\citep{kawaguchi2015Bayesian} as acquisition function.

\begin{figure}[t]
  \centering
  \subfloat[]{
    \includegraphics[width=0.5\linewidth]{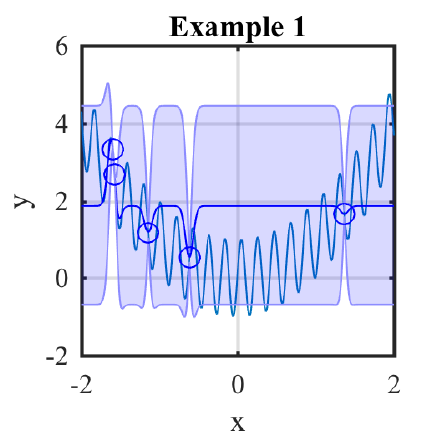}%
  }
  \subfloat[]{
    \includegraphics[width=.5\linewidth]{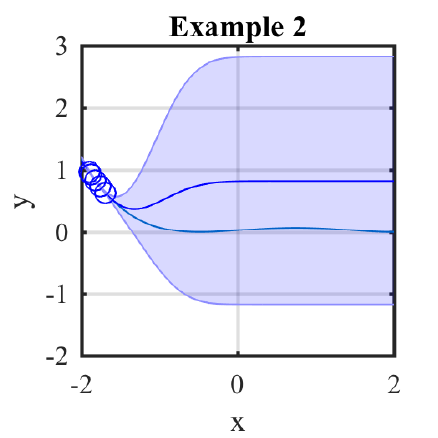}%
  }
  \caption{\label{fig:drawbacks} Illustration for mis-leading online
    length-scale selection for the SE kernel \eqref{eq:k_SE} based on
    max-likelihood (left) and poor prediction of the minimum using the
    SE kernel
    (right).  GP-mean (bold
    blue line), GP-variance (shaded blue area) and true underlaying
    function (turquoise). Such misleading ML estimates of
    hyperparameters may be due to a randomized set of initial
    observations or if the function contains a low and high frequency
    part and we directly jump into high frequency modelling due to a
    small number of samples.}
\end{figure}

In this paper we rather focus on the first question, the choice of
model or prior over the objective function. Clearly, from the purely
Bayesian stance the prior must be given and is not subject to
discussion. However, there are a number of reasons to reconsider this:

\emph{Choice of Hyperparameters:} A large number of methods, notably
with the exception of \cite{hernandez2014predictive}, rely on an
online point estimate of the hyperparameters. Many convergence proofs
in fact rely on an apriori chosen GP hyperprior (see
also Tab.~\ref{tab:acq_fcn} for
details). Several experimental results are reported where hyperpriors
have been carefully chosen by hand or optimized on samples from the
objective function which are apriori not given
\citep{srinivas2012information, wang2014bayesian,
  kawaguchi2015Bayesian}.  In practise, choosing the hyperprior online
(e.g.\ using leave-one-out cross-validation (LOO-CV) on the so-far
seen data) is prone to local optima and may lead to significant
inefficiency w.r.t.\ the optimization process. For instance, in
Fig.~\ref{fig:drawbacks}(a) we see that with a maximum likelihood
estimate of the length-scale we may get an unuseful view of the
objective function. Taken together, we believe that online selection
of hyperparameters during online learning remains a key challenge.

In this paper we take the stance that if one chooses a point estimate
for the hyperprior online, then maximum likelihood only on the seen
data is \emph{not} an appropriate model selection criterion. Instead,
we should choose the hyperprior so as to accelerate the optimization
process. We propose to \emph{cool down} a length-scale parameter based
on hyperprior selection \emph{w.r.t.\ the acquisition function}, that
is, which hyperprior promises more ``acquisition'' with the next query
point. We combine this with a lower-bound heuristic for robustness.

\emph{Choice of kernel function:} The squared-exponential kernel is
the standard choice of prior. However, this is in fact a rather strong
prior as many relevant functions are heteroscedastic (have different
length-scales in different regions) and have various local optima,
each with different non-isotropic conditioning of the Hessian at the
local optimum. Only very few preliminary experiments on
heteroscedastic and non-isotropic models have been reported
\citep{MartinezCantin15nipsws,mohammadi2016small}.

In this paper we propose a novel type of kernel function with the
following in mind. Classical model-based optimization of convex
black-box functions \citep[Section 8]{nocedal2006numerical}
is extremely efficient \emph{iff} we know the function to be
convex. Therefore, for the purpose of optimization we may presume that
the objective function has local convex polynomial regions, that is,
regions in which the objective function is convex and can reasonably
be approximated with a (non-isotropic) 2nd-order polynomial, such that
within these regions, quasi-Newton type methods converge very
efficiently. Like it would be the case in Fig \ref{fig:drawbacks}(b).
To this effect we propose the Mixed-Global-Local (MGL) kernel,
which expresses the prior assumption about local convex polynomial
regions, as well as automatically implying a local search strategy
that is analogous to local model-based optimization. Effectively, this
choice of kernel integrates the efficiency of local model-based
optimization within the Bayesian optimization framework.

In summary, our contributions are
\begin{itemize}
\item An online hyperparameter cool down method based on the acquisition
  function instead of maximum likelihood, which we call alpha-ratio
  cool down, and a length-scale lower bound for robustness.
\item A novel kernel function to represent local convex polynomial
  regions, that implies local quadratic interpolation and
  optimization steps analogous to classical (quasi-Newton-type)
  model-based optimization combined with global Bayesian optimization.
\item An efficient algorithm for detecting local convex polynomial regions.
\end{itemize}

Our paper is structured as follows: After giving the essential BO
background, we explain a length-scale cool down scheme for isotropic
kernels. Next we introduce the MGL kernel which will make use of the
adaption scheme from the first step.  In the end we will show
significant performance improvements compared to classical BO with
'optimal' (see Rem.~\ref{Rem:optimalParameters}) hyperparameters.

\subsection{Problem Statement and Bayesian optimization background}
\label{Subsec:ProblemBackground}

We consider the black-box optimization problem
\begin{equation}\label{eq:generalMinProblem}
	\bm x^*=\argmin_{\bm x\in\DD}f(\bm x)
\end{equation}
with an objective $f:\DD \rightarrow \mathbb R$ that maps a
hypercube
\begin{equation}\label{eq:generalHypercube}
\DD=\{ \bm x\in\mathbb{R}^d\mid x_i \in [0,\;1]\subset\mathbb{R}, i=1,2,..,d\}
\end{equation}
to real numbers. We consider a Gaussian
process (GP) \citep{rasmussen2006gaussian} prior over $f$ with
constant prior mean function $\mu = c_\mu$. Together with a
covariance (kernel) function $k(\bm x, \bm x')$ we write
$\mathcal{GP}(c_\mu,k)$ for the prior GP in short.  Given samples $\bm
y_N=[y_1,..,y_N]^T$ at points $X_N=[\bm x_1,..,\bm x_N]^T$ where
$y_n=f(\bm x_n)$, the posterior GP reads
\begin{align}
\mu_N(\bm x) &= \bm c_\mu + \bm k_N(\bm x)^T {K_N}^{-1}(\bm y_N-\bm c_\mu)\label{eq:GP_mu}\\
k_N(\bm x,\bm x')&=k(\bm x,\bm x')-\bm k_N(\bm x)^T {K_N}^{-1}\bm k_N(\bm x')\\
\sigma_N^{2}(\bm x) &= k_N(\bm x, \bm x).\label{eq:GP_sigma}
\end{align}
$\bm k_N(\bm x)=[k(\bm x_1,\bm x),..,k(\bm x_N,\bm x)]^T$,~
$k_N(\bm x,\bm x')$ is the posterior covariance,
$K_N$ is the positive definite kernel or
matrix $[k(\bm x,\bm x')]_{\bm x,\bm x'\in X_N}$ and
$\bm c_\mu$ the prior mean, stacked in a vector.
A very common choice of kernel  is the squared exponential (SE) kernel
\begin{align}\label{eq:k_SE}
	k_{\text{SE}}(\bm x,  \bm x') = \sigma_f^2
		  \exp\left(- 0.5 \frac{||\bm x - \bm x'||^2}{l^2}
		  	   \right)
\end{align}
with hyperparameters $l$ (length-scale) and $\sigma_f^2$
(prior variance).
The GP model assumption about $f$ builds the basis for many
BO algorithms.
A general prototype for such an algorithm is given in
Alg.~\ref{alg:generic_Bayesian_optimization}, where $\alpha_t$
represents the algorithm specific acquisition function.
For experiments we will use the well known and theoretical
extensively
studied Expected Improvement (EI) \citep{bull2011convergence}
acquisition function, which is defined as
\begin{align}\label{eq:EI_acquisition function}
	\alpha_{\text{EI}}(\bm x) = 
	\begin{cases}
		-\sigma_{n-1}(\bm x)(u\Phi(u) + \phi(u)),\; \sigma_{n-1}(\bm x) > 0\\
		\min(f^*_{n-1}-\mu_{n-1}(\bm x),0), \; \sigma_{n-1}(\bm x) = 0
	\end{cases}
\end{align}
with $u = \frac{f^*_{n-1}-\mu_{n-1}(\bm x)}{\sigma_{n-1}(\bm x)}$
and $f^*_{n-1} = \min_i \bm y_{n-1,i}$.

\begin{table}[t]
\caption{Overview of common BO algorithms and their
properties: Performance guarantees (T) and
online model adaption (A). Only in case of EI the model adaption is
explicitly treated
in performance analysis.} \label{tab:acq_fcn}
\begin{center}
\begin{tabular}{lccc}
ACQUISITION FCN $\alpha_t$& T & A \\
\hline \\
Expected Improvement (EI) & \cmark & \cmark\\
~\citep{bull2011convergence} &&\\
Upper Confidence Bound (UCB) & \cmark & \xmark \\
~\citep{srinivas2012information} &&\\
Entropy Search (PES) & \xmark & \xmark \\
~\citep{hernandez2014predictive} &&\\
Exponential Convergence & \cmark & \xmark \\
~\citep{kawaguchi2015Bayesian} &&
\end{tabular}
\end{center}
\end{table}

\subsection{Related work}

Classical online model adaption in BO
through maximum likelihood or LOO-CV \citep{rasmussen2006gaussian}
are compared in \cite{bachoc2013cross}. LOO-CV
turns out to be more robust under model misspecification whereas
maximum likelihood gains better results as long as the model is chosen
'well'. \cite{jones2001taxonomy} already discusses the problem of highly
misleading initial objective function samples.
In \cite{forrester2008global} an idea earlier mentioned in
\cite{jones2001taxonomy} is studied, where the hyperparameter
adaption is combined with the EI calculation
in one step. The authors show better performance in certain cases,
while the resulting sub-optimization can get quite tedious as
described in \cite{quttineh2009implementation}.
Another approach for improving the length-scale hyperparameter
adaption is presented in \cite{wang2016Bayesian}. They try
to limit local exploration by setting an upper bound on the
length-scale. The bound they propose is independent of dimensionality
and experimentally chosen by hand for their applications.
Those papers address the same problem
but all of them are modifications of the basic approach to adjust
hyperparameters to maximize data likelihood.
In contrast we adjust hyperparameters based on the aquisition
function, aiming at optimization performance rather than data likelihood.

\cite{mohammadi2016small} introduce the idea of \emph{local}
length-scale adaption based on maximizing the acquisition function
(EI) value, which is not efficient as they say (and different to the
cool down we propose). Nevertheless we endorse the underlying idea,
since it is related to our motivation.

On the model side there are several ideas which yield non-isotropic
models by building an ensemble of local isotropic kernels, e.g.\ based
on trees \citep{assael2014heteroscedastic}.  We however introduce a
specific kernel rather than a concept of combining kernels or Gaussian
processes taylored for improving BO.

There are also concepts regarding locally defined kernels, e.g.\
\cite{krause2007nonmyopic}.
The idea of \cite{MartinezCantin15nipsws} is somehow
closely related to ours, because they use a local and a global kernel function,
which is a great approach, as we believe.
They parametrize the location of the local kernel as well
as the respective parameters. Consequently they
end up with a large number of hyperparameters which makes
model selection very difficult.
In constrast to their work we are able to gain comparable or better
performance in well-known benchmarks. At the same time
we overcome the problem of many hyperparameters
by a separated, efficient algorithm for determining the
location of local minimum regions. Furthermore we use a non-isotropic
kernel for better fitting local minimum regions.

As a last aspect we want to mention, that PES, \citep{hernandez2014predictive}
also incorporates a local minimum region criterion encapsulated in the
acquisition function itself, what we find is very
interesting and is
conceptually connected to our ideas. We compare the
performance of PES
along with another state-of-the-art BO algorithm
\citep{kawaguchi2015Bayesian} against our
advances in combination with classical EI
in the experimental section.

\begin{algorithm}[t!]
\caption{General Bayesian optimization}\label{alg:generic_Bayesian_optimization}
\begin{algorithmic}[1]
\Procedure{GBO}{objective $f$, $\mathcal{GP}(c_{\mu},k)$
	, max. Iterations $N$, acquisition function $\alpha$}
	\State init $X_0 =\{\bm x_{01},...,\bm x_{0N_i}\}$, $\bm x_0\in\mathcal{D}$
	\State init $\bm y_0 = [f(\bm x_{01}),...,f(\bm x_{0N_i})]^T$
	\State $n\gets1$
	\For{$n\leq N$}
		\State perform model adaption with $\lbrace X_n,\bm y_n\rbrace$
		\State $\bm x_n=\argmin_{\bm x \in \mathcal{D}} \alpha_n(\bm x)$
			\Comment \textit{acq. function val.}
		\State $X_n\gets\lbrace\bm x_n\rbrace \cup X_{n-1}$
		\State $\bm y_n\gets \lbrace f(\bm x_n)\rbrace \cup \bm y_{n-1}$
			\Comment \textit{query new sample}
		\State $n = n + 1$
	\EndFor
	\State $n^* \gets \argmin_n y_n \in \bm y_N$
	\State \textbf{return} $\bm x_{n^*}$
		\Comment \textit{best observation}
\EndProcedure
\end{algorithmic}
\end{algorithm}


\section{LENGTH-SCALE COOL DOWN}\label{section:correlationCoolDown}

In this section we address length-scale
adjustment of an isotropic kernel
during the optimization process as part of the general BO Algorithm
(Alg.~\ref{alg:generic_Bayesian_optimization}, Line 6).
Most commonly, $l_n$ at iteration $n$ is chosen such that the model
represents the observed data $\lbrace X_n,\bm y_n\rbrace$ well.  This
approach however ignores the closed loop, i.e., that the next sample
choice will depend on $l_n$. In contrast we aim to control the
optimization process characteristics by performing a transition from
``macro'' to ``micro'' modelling of the objective function.

\subsection{Alpha-ratio cool down}

Let $l_{n-1}$ be the length-scale used in the previous iteration. In
our approach we want to decide whether to reuse the
same length-scale or decrease it to a specific smaller length-scale
$\tilde l_n < l_{n-1}$ in iteration $n$. In our experiments we will
choose $\tilde l_n = \max(l_{n-1}/2, \bar l_{n})$, where $\bar l_{n}$
is a hard lower bound we present in the next section. Neglecting this
bound, we want to decide whether to half the length-scale.

We propose to use the aquisition function as a criterion for this
decision. Let
\begin{align}\label{eq:alphaRatio}
	\alpha_{r,n} := \frac{\alpha^*(\tilde l_n)}{\alpha^*(l_{n-1})}
\end{align}
be the alpha-ratio, where $\alpha^*(l) = \min_{\bm x\in\DD} \alpha_n(\bm x;l)$ is the
optimal aquisition value when using length-scale $l$. In typical
situations we expect that $\alpha_{r,n} >1$ because the reduced
length-scale $\tilde l_n$ leads to larger posterior variance, which
typically leads to larger aquisition values, i.e., more chances
for progress in the optimization process. We turn this argument
around: if $\alpha_{r,n}$ is not \emph{substantially} larger than 1,
then choosing the smaller length-scale $\tilde l_n$ does not yield
substantially more chances for progress in the optimization
process. In this case, as a smaller length-scale has higher risk of
overfitting (Fig.~1(a)), we decide to stick to the old length-scale
$l_{n-1}$.

In summary, in our \textit{alpha-ratio (AR) cool down} for
length-scale adaption we have a fixed threshold $\bar \alpha_{r}>1$
and choose $l_n=\tilde l_n$ as new length-scale if $\alpha_{r,n}>\bar
\alpha_{r}$, and $l_n=l_{n-1}$ otherwise. In
Alg.~\ref{alg:alphaRatioAdaption} we summarize the scheme, where in lines 2-5
the calculation of the alpha-ratio $\alpha_{r,n}$ is shown.

\begin{remark}
It is straight-forward to transfer existing convergence guarantees of
Bayesian Optimization, as given in \cite{bull2011convergence}, to hold
also in the case of AR cool down by a constant absolute lower bound
$\bar l$, such that we have
\begin{align}
	l \in [\bar l, l_1]\subset \RR^+
\end{align}
throughout the optimization.
\end{remark}

\begin{algorithm}[t]
\caption{Alpha-ratio (AR) cool down}\label{alg:alphaRatioAdaption}
\begin{algorithmic}[1]
\Procedure{AR}{$\mathcal{GP}(0,k)$, acquisition function $\alpha(.)$,
	$\alpha_{n-1}^*$, $\bar \alpha_r$, $l_{n-1}$, $\bar c$,
	iteration $n$, dimension $d$}
	\State calculate $\bar l_{n}(d,\bar c)$ \Comment \textit{lower bound see
		\eqref{eq:lowerBound}}
	\State $\tilde l_n \gets \max\lbrace l_{n-1}/2,\bar l_{n}(d,\bar c) \rbrace$
	\State $\alpha^*(\tilde l_n)\gets\min_{\bm x \in \mathcal{D}} \alpha_n(\bm x;\tilde l_n)$
		\Comment \textit{acq. with $\tilde l_n$}
	\State $\alpha_{r,n}\gets \alpha^*(\tilde l_n)/\alpha^*_{n-1}$
	\If{$\alpha_{r,n}>\bar \alpha_r$}
		\State $l_n \gets \tilde l_n$ \Comment \textit{reduce length-scale}
	\Else
		\State $l_n \gets l_{n-1}$
			\Comment{keep length-scale}
	\EndIf
	\State \textbf{return} $l_n$
\EndProcedure
\end{algorithmic}
\end{algorithm}

\subsection{Lower Bound on Correlation}

To increase the robustness of length-scale adaptation especially in
the very early phase of optimization
  we propose a
length-scale lower bound $\bar l_n$ that explicitly takes the search
space dimensionality into account. To find such a lower bound, we
assume that we have an idea about minimum correlation $\bar c\in \RR$
that two points $\bm x,\bm x'$ should have at all time.  For
constructing the lower bound on length-scale we consider the following
best case regarding the correlation: The samples $\lbrace\bar
X_n,\bar{\bm y}_n\rbrace$ are chosen in a way, such that the minimal
correlation of any new point $\bm x \in \DD$ with the data is highest. Based on $\bar
X_n$ for different number of samples $n=|\bar X_n|$, we calculate the
corresponding minimum length-scale $\bar l_n$ to satisfy the minimum
required correlation $\bar c$. Since it is very likely, that samples
acquired by Alg.~\ref{alg:generic_Bayesian_optimization} will violate
the best case assumption, the calculated length-scale serves as a
lower bound. Formally consider the set
\begin{align}\label{eq:worstCaseSet}
	\bar X_n :=
	\argmax_{X, \text{ s.t. } |X| = n,\bm x' \in X} \left(
		\min_{\bm x\in \DD} k(||\bm x-\bm x'||;l)\right)
\end{align}
and $k(.;l)\in (0,1]$ (for the design scheme)
which yields highest correlation values for an ``adversarial'' $\bm x \in \DD$.
Note that $l$ in $k(.;l)$
is not relevant for \eqref{eq:worstCaseSet}. Even if an
acquisition function will not explore the search space in terms of
\eqref{eq:worstCaseSet} we at least want to guarantee
that in case of \eqref{eq:worstCaseSet} we would have
a minimum correlation. To this end we solve
\begin{align}\label{eq:solveForlength-scale}
	\forall \bm x, \bm x' \in \bar X_n:
		k(||\bm {\bar x} - \bm {\bar x'}||;\bar l_n) \overset{!}{\geq} \bar c
\end{align}
for $\bar l_n$, which is the length-scale lower bound we are searching for.

\begin{remark}\label{Rem:1D_best_case}
Note that \eqref{eq:worstCaseSet} is NP-hard in the sense
of the sphere packing problem for dimensionality $d>1$. But
for $d=1$ the solution to \eqref{eq:worstCaseSet} is
\begin{equation}\label{eq:fullGrid1D}
	\bar X_{n-1} = \lbrace 1/n, 2/n, ..,(n-1)/n \rbrace.
\end{equation}
\end{remark}

\subsubsection{Approximation for higher dimensions}
	\label{Subsubsection:Approximationlength-scaleLowerBound}

Now we use the 1D result (Rem. \ref{Rem:1D_best_case})
with the following idea in order to transfer
the bound to higher dimensions: The max. distance $\delta_{1,n}$ in 1D between
samples implies a minimal, empty (sample free) space. We demand that in
higher dimensions the ratio between the volume of the whole search space
($1^d = 1$) and this minimal empty space remains the same. Thereby we seek to
get equal coverage of every sample in case $d>1$.
Following this idea, we transform the minimum distance
\begin{align}
	\delta_{d,n} :=\min_{\bm x, \bm x' \in \bar X_n} ||\bm x-\bm x'||
\end{align}
for $d=1$ ($\delta_{1,n}=1/n$, compare with Rem. \ref{Rem:1D_best_case}) in terms of the sphere volume
\begin{align}\label{eq:volume}
	V_d(r) = \frac{\pi^{d/2}r^d}{\Gamma(\frac{d}{2} +1)}
\end{align}
\citep{weisstein},
to obtain a reasonable approximation for higher dimensions $d>1$. I.e. given
$\delta_{1,n}$ we calculate $V_1(\delta_{1,n})$
and solve
\begin{align}
	V_d(\delta_{d,n}) \overset{!}{=}  V_1(\delta_{1,n})
\end{align}
for $\delta_{d,n}$
which allows us together with $||\bar{\bm x}-\bar{\bm x}'||\approx \delta_{d,n}$
as in
\eqref{eq:solveForlength-scale} to obtain the corresponding
length-scale lower bound. By applying these steps using the
example of the SE kernel \eqref{eq:k_SE} we get
\begin{align}\label{eq:lowerBound}
	\bar l_{n}(d,\bar c) = \sqrt{-\frac{1}{2\log(\bar c)}}
		\left(
			\frac{\Gamma(\frac{d}{2}+1)}{\Gamma(\frac{3}{2})}
			\pi^{0.5(1-d)}
			\underbrace{
				\frac{1}{n}
				}_{
				\delta_{1,n}
				}
		\right)^{\frac{1}{d}}
\end{align}
as lower bound for the length-scale by setting $\sigma_f = 1$.

\section{MIXED-GLOBAL-LOCAL KERNEL}\label{section:localQuadKernel}

In this section we introduce the MGL kernel.
We assume that each local (global) optimum $\bm x^*_i$ of
(\ref{eq:generalMinProblem})
is within a neighbourhood $\UU_i(\bm x^*_i)$ that can be
approximated by a positive definite quadratic function.
More precisely:
\begin{definition}\label{def:local_neighbourhood}
  Given a data set $D=\{(\bm x_i,y_i)\}$, we call a convex subset
  $\UU\subset\DD$ a convex neighborhood if
        the solution of the regression problem
 	\begin{align}\nonumber
		\lbrace \beta_0^*, \bm \beta_1^*, B^*\rbrace =
		\\\label{eq:regression}
		\argmin_{\beta_0, \bm \beta_1,B}
			\sum_{k:\bm x_k\in\UU}
			\left[( \beta_0 + \bm \beta_1^T\bm x_k + \frac{1}{2}\bm x_k^TB\bm x_k)
			- y_k\right]^2 ~,
	\end{align}
	($\bm x_k \in \UU$ the data points in $\UU$) has a positive definite Hessian $B$.
\end{definition}

If we are given a set $\{\UU_i\}$ of convex neighborhoods that are
pair-wise disjoint we define the following kernel function:
\begin{definition}
The Mixed-Global-Local (MGL) kernel is given by
\begin{align}\label{eq:k_MGL}
	k_{\text{MGL}}(\bm x, \bm x') =
	\begin{cases}
		k_q(\bm x, \bm x'),
			\; \bm x, \bm x' \in \UU_i,\\
		k_s(\bm x, \bm x'), \bm x \notin \UU_i, \bm x' \notin \UU_j\\
		0, \; \text{else}
	\end{cases}
\end{align}
for any $i,j$,

where $k_s$ is a stationary-isotropic kernel \citep{rasmussen2006gaussian}

and
\begin{align}\label{eq:k_q}
	k_q(\bm x, \bm x') = (\bm x^T \bm x'+1)^2
\end{align}
the quadratic kernel.
\end{definition}
This kernel is heteroscedastic in the sense that the quadratic kernels
in the convex neighborhood implies fully different variances than the
``global'' stationary-isotropic kernel around the neighborhoods.
Due to the strict seperation of the corresponding regions,
the posterior calculation can be decoupled.

\subsection{MGL kernel hyperparameter adaption}

The hyperparameters of the MGL kernel are the length-scale and other
parameters (prior mean $c_\mu$, prior variance $\sigma_f$) of the
isotropic kernel $k_s$ and the convex neighborhoods $\UU_i$.  For the
model update in Alg.~\ref{alg:generic_Bayesian_optimization}, line 6
we have to update these hyperparameters. For the length-scale we use
Alg.~\ref{alg:alphaRatioAdaption} and adapt the remaining parameters
with maximum likelihood or LOO-CV. For determining $\UU_i$ we
introduce Alg.~\ref{alg:locMinIdent}. The main simplification is the
discretization of the search space using the samples as centers for
k-nearest-neighbor (kNN) search. As soon as a kNN tuple of samples
satisfy Def.~\ref{def:local_neighbourhood}, we get a ball shaped
local minimum region. In line 6,7 we calculate a potential local
minimum region. In line 8 we fit a quadratic form using the samples
inside this region. Lines 9-27 are used as selection criterion for
local regions. Besides trivial criterions, in line 21 we add a local
convergence criteria.  In line 25 we force close points to the local
region to contribute to the local model which turned out to improve
overall performance.  Line 31 removes all regions that overlap with
better regions in order to find a disjoint set of convex neighborhoods.
\begin{remark}
	Let $\bar \UU$ be a real local minimum region.
	We will have at most $Vol(\bar \UU)/\epsilon^d<\infty$ detected
	regions inside $\bar \UU$ due to Alg.~\ref{alg:locMinIdent}, line 21,
	and the fact that the search space \eqref{eq:generalHypercube} is compact.
	Therefore the asymptotic
	behavior for $n\rightarrow\infty$ will not change for
	any acquisition function, since we have $k_{MGL}\rightarrow k_s$
	in this case.
\end{remark}

\begin{algorithm}[t]
\caption{Local minimum region identification}\label{alg:locMinIdent}
\begin{algorithmic}[1]
\Procedure{LMRI}{samples $\lbrace X, \bm y \rbrace$, dimension $d$,
	abs. tolerance $\epsilon$, }
	\State $N_R \gets 0$
	\Comment{\# local regions}
	\For{$i = 1,2,..,|\bm y|$}
		\State $N_U\gets
			1+d+\frac{d(d+1)}{2}$
		\Comment{\# unknowns}
		\For{ $k=N_U,N_U+1..\min(2N_U,|\bm y|-1)$}
			\State $X_{\text{kNN}} \gets kNN \text{ of } X_{i,1:d} \text{ in }X$
			\State $\UU_{N_R} \gets $ calculate NN ball of $X_{\text{kNN}}$
			\State $\lbrace \beta_0^*, \bm \beta_1^*, B^*\rbrace \gets$regression ($X_{\text{kNN}}, \bm y_{\text{kNN}}$)
			\If{ $B^*$ is not pos. definite }
				\State break
				\Comment \textit{no local minimum region}
			\EndIf
			\State $\bm x_i^* \gets -0.5B^{*-1}\bm \beta_1^*$ \Comment local optimum
			\State $y_i^* \gets \beta_0^* + \bm \beta_1^* \bm x_i^* +
				\bm x_i^{*,T}B^*\bm x_i^*$
			\If{ $\bm x_i^* \notin \UU_{N_R}$}
				\State break
				\Comment \textit{min. outside of local region}
			\EndIf
			\If{ $ y_i^* > \min(\bm y)$}
				\State break
				\Comment \textit{no improvement}
			\EndIf
			\State $\bm x_{1NN}\gets 1NN \text{ of } \bm x_i^* \text{ in }X$
			\If{ $||\bm x_{1NN} - \bm x_i^*)||<\epsilon$}
				\State break
				\Comment \textit{local convergence}
			\EndIf
			\State $\tau \gets \min_{\bm x \in X\backslash X_{kNN}}||\bm x-\UU||$
			\If{ $\tau < 0.05$}
				\State break
				\Comment \textit{ignorance of neighborhood}
			\EndIf
			\State $N_R \gets N_R + 1$
		\EndFor
	\EndFor
	\State remove overlapping $\UU_i$, highest (worst) $y_i^*$ first
	\State \textbf{return} all $\UU_i$
		\Comment \textit{local minimum regions}
\EndProcedure
\end{algorithmic}
\end{algorithm}



\section{EMPIRICAL RESULTS}\label{section:results}

We first give some illustrative examples of the behavior of the
alpha-ratio cool down and the MGL kernel, before comparing
quantitatively the performance against other methods on benchmark
problems.

The original source code which was used to generate all of
the results can be found in the supplementary material and
will be published.
For all tests we choose the following configurations: For
Alg.~\ref{alg:alphaRatioAdaption} we set $\bar c=0.2$, $\bar
\alpha_r=1.5$ and for Alg.~\ref{alg:locMinIdent} we choose
$\epsilon=1e-9$.  For the MGL-kernel \eqref{eq:k_MGL} we take the SE
kernel \eqref{eq:k_SE} for $k_s$.  We estimated the observation
variance $\sigma_f^2$ in \eqref{eq:k_SE} and the constant mean of the
prior GP via maximum likelihood and scaled the observation variance
down by factor $100$ for consistency with the quadratic part of
\eqref{eq:k_MGL} if any local region is detected. For computing
Alg.~\ref{alg:generic_Bayesian_optimization} line 7 we first solved
the minimization using the $k_s$ kernel of \eqref{eq:k_MGL} and
compared it with the results of the minimization problems using the
$k_q$ kernel for each local minimum region $\UU_i$ since all the
regions for the different kernel parts are disjoint.  We used three
samples as initial design set, chosen by latin hypercube sampling.

\begin{remark}\label{Rem:optimalParameters}
In the following we will often refer to an \emph{optimal} choice of
hyperparameters. By this we mean that 1000 random samples from the
respective objective function are taken. On this data an exhaustive
LOO-CV is used to select the length-scale, and max-likelihood to select
the prior variance $\sigma_f^2$ and mean-prior $c_\mu$.
\end{remark}

\begin{figure*}[t!]
\centering
	\subfloat[Immediate regret (IR) for optimizing a GP sample function with
	zero mean, unit process noise and SE kernel function with
        $l=0.1$.
	Shown are
	online LOO-CV and AR length-scale adaption vs.\ optimal (Rem.~\ref{Rem:optimalParameters})
        length-scale. See Sec.~\ref{secExp1} for details.]{
  		\includegraphics[width=2.0in]{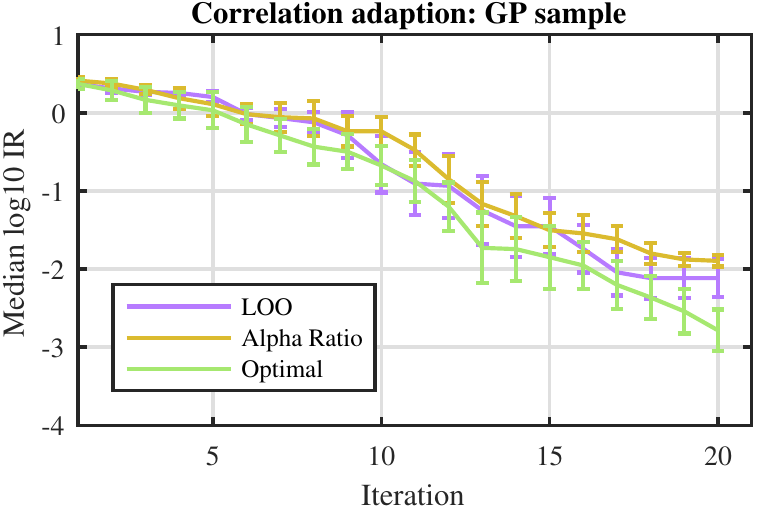}%
	}\quad
	\subfloat[Plot of a counter example objective function for
	ordinary model adaption: GP with zero mean, unit observation variance and
	SE kernel function with $l=0.05$ with
	local quadratic regions inserted.]{
  		\includegraphics[width=2.0in]{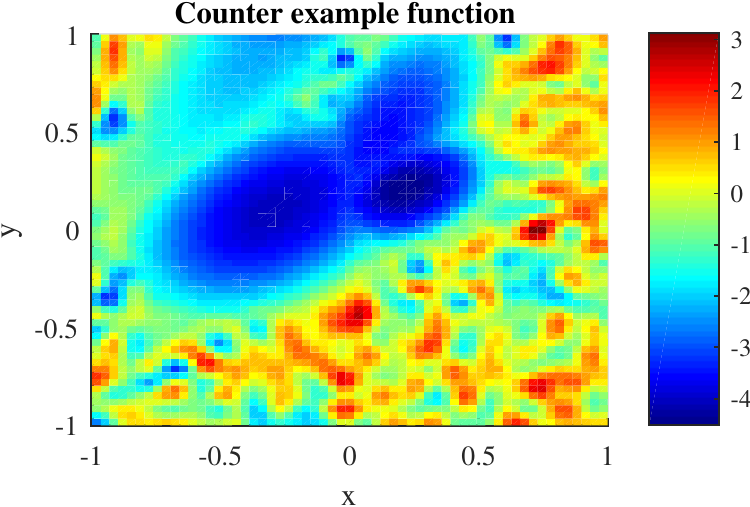}%
	}\quad
	\subfloat[Immediate regret (IR) in case of optimizing the objective
	function in Fig.~\ref{fig:ExamplesLenthscaleAdaption}(b) with
	online LOO-CV and AR length-scale adaption vs.\ optimal (Rem.~\ref{Rem:optimalParameters})
	length-scale. See Sec.~\ref{secExp1} for details.]{
  		\includegraphics[width=2.0in]{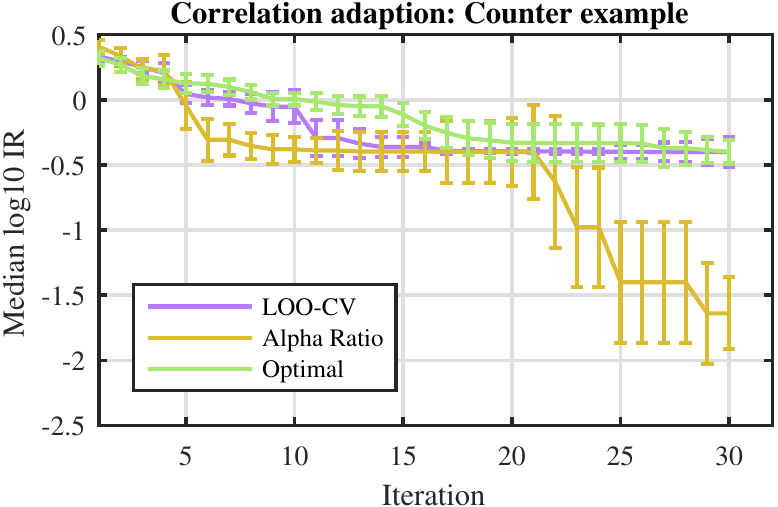}%
	}\\
	\subfloat[Cool down of length-scale by controlling the
	alpha-ratio \eqref{eq:alphaRatio}. Shown is the ratio
	and the reduction votes by the alpha-ratio cool down, indicated by the
	bars until value of one.]{
  		\includegraphics[width=2.0in]{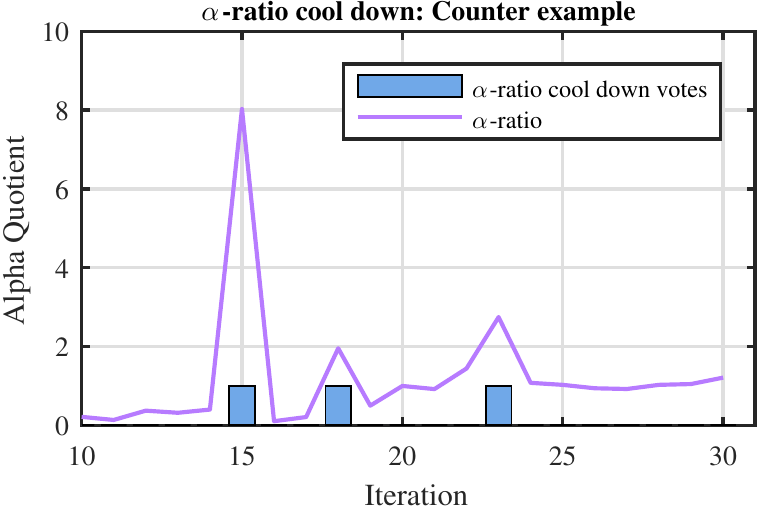}%

  	}
  	\quad
  	\subfloat[Plot of an example objective with local quadratic regions:
	GP with zero mean, unit observation variance and
	SE kernel function with $l=0.3$ with
	local quadratic regions inserted.]{
  		\includegraphics[width=2.0in]{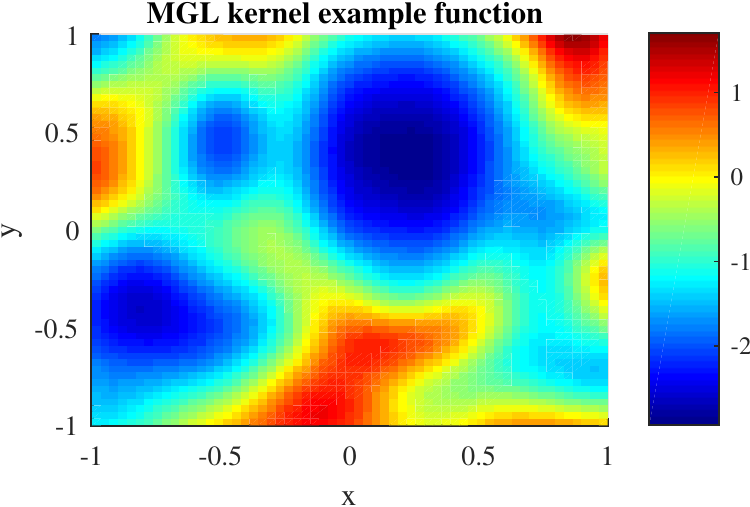}%
	}\quad
	\subfloat[Immediate regret (IR) for optimizing the objective
	function in Fig.~\ref{fig:ExamplesLenthscaleAdaption} (e) with
	AR length-scale adaption and optimal length-scale vs.\
	AR adaption and the MGL kernel \eqref{eq:k_MGL}.]{
  		\includegraphics[width=2.0in]{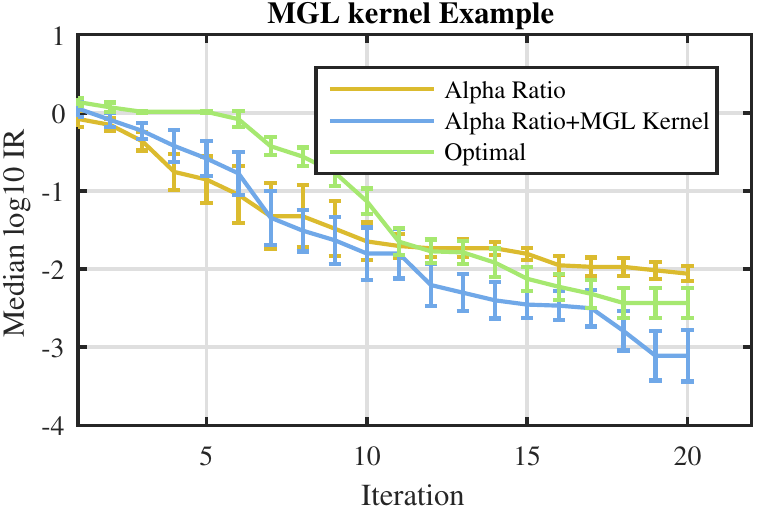}%
  	}
\vspace{.2in}
\caption{Illustrations of the performance and drawbacks of ordinary
model adaption, our alpha-ratio cool down and MGL kernel.
EI is used as acquisition function.}
\label{fig:ExamplesLenthscaleAdaption}
\end{figure*}

\subsection{Illustrating the AR cool down}\label{secExp1}

To illustrate AR cool down (Alg.~\ref{alg:alphaRatioAdaption}) within
Bayesian Optimization (Alg.~\ref{alg:generic_Bayesian_optimization})
we consider the ideal setting where the objective $f$ is sampled from
a GP with a stationary SE kernel. In
Fig.~\ref{fig:ExamplesLenthscaleAdaption}(a) we display the regret
when using 1) the \emph{optimal} parameters
(Rem.~\ref{Rem:optimalParameters}), 2) online

 an exhaustive \emph{LOO-CV} for length-scale
parameter and $\sigma_f^2$ and $c_\mu$ estimated from data using maximum likelihood,
and 3) our
AR cool down method. As the true objective is indeed a sample from a
stationary isotropic GP, online LOO-CV and optimal hyperparameters work well,
while AR cool down has less variance in performance.
bel{fig:ExamplesMINkernel}

To give more insight, in Fig.~\ref{fig:ExamplesLenthscaleAdaption}(d)
the mechanics of Alg.~\ref{alg:alphaRatioAdaption}
are illustrated. Shown is the alpha-ratio \eqref{eq:alphaRatio}
and the cool down events are indicated by the small bars.

Finally, we illustrate AR cool down in case of the non-stationary
heteroscedastic objective function given in
Fig.~\ref{fig:ExamplesLenthscaleAdaption}(b). Note its structure
mixing small and large length-scale local optima. The regret curves in
Fig.~\ref{fig:ExamplesLenthscaleAdaption}(c) compare the same three
methods described above. Both, the optimal and online LOO-CV methods try to fit
stationary GP hyperparameters to a heteroscedastic objective
function. Both of these lead to rather poor optimization behavior. AR
cool down behaviors clearly superior.


\subsection{Illustrating the MGL kernel}

Fig.~\ref{fig:ExamplesLenthscaleAdaption}(e) displays another
objective function that corresponds to the MGL kernel assumption. As
shown in Fig.~\ref{fig:ExamplesLenthscaleAdaption}(f), using the MGL
kernel together with the AR cool down we are able to gain a
performance improvement compared to the plain AR cool down or even the
optimal hyperparameters.



\subsection{Comparision with PES, IMGPO and EI}

\begin{figure*}[t!]
\centering
	\includegraphics[width=\linewidth]{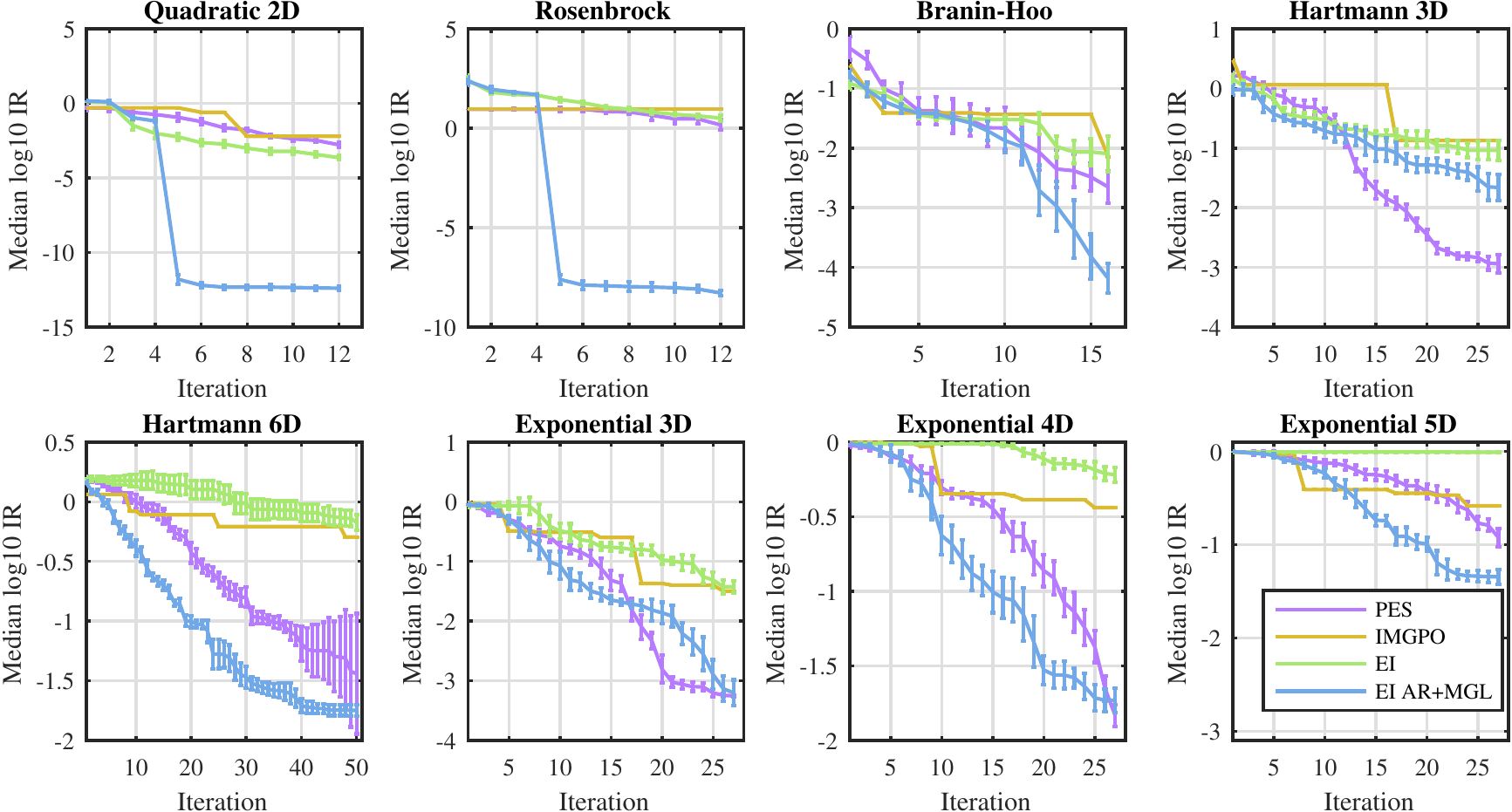}
	\vspace{.2in}
	\caption{Comparison of recent Bayesian optimization
		algorithms with synthetic test
		functions. See Rem.~\ref{Rem:optimalParameters}.}
	\label{fig:empiricalResults}
\end{figure*}

In Fig.~\ref{fig:empiricalResults} we report on results using several
synthetic benchmark functions.  Shown are predictive entropy search
(PES) \citep{hernandez2014predictive} (which treats hyperparameters in
a Bayesian way in the acquisition function), infinite metric GP optimization (IMGPO)
(which uses a Bayesian update for hyperparameters in each iteration),
classical EI with
optimal (Rem.~\ref{Rem:optimalParameters}) hyperparameters, and EI
using our alpha-ratio model adaption and the MGL-kernel (EI AR + MGL).

For all performance tests where we show the log10 median
performance, we made 32 runs and estimated the median variance
via bootstrapping. The errorbars indicate one times the
standard deviation.

The results of
PES\footnote{\texttt{https://github.com/HIPS/Spearmint/tree/PESC}} and
IMGPO\footnote{\texttt{http://lis.csail.mit.edu/code/imgpo.html}}
search are obtained using the code available online. Therefore the
results are somehow biased through the prior on the hyperparameters
in these algorithms.

In addition to commonly considered benchmark functions
(Rosenbrock, Branin-Hoo, Hartmann3D, Hartmann 6D)
taken from \cite{simulationlib}, we show
a simple quadratic function in the interval
$[-2,2]^2$ and an exponential function
of the form
$f_{\text{exp}}(\bm x) = 1-\exp(\bm x^TC\bm x)$ with
$C:=\text{diag}([10^{0/(d-1)},10^{1/(d-1)},..,10^{(d-1)/(d-1)}])$ on
the same interval in respective dimensions $d$.

The MGL-kernel outperforms significantly in case of the quadratic and
the more quadratic like Rosenbrock objective. Also for Branin-Hoo,
Hartmann 6D and Exponential 5D our method significantly outperforms
existing state-of-the-art Bayesian optimization methods. In case of
Hartmann 3D, PES turns out to work better. Nevertheless we want to
emphasize the outstanding improvement compared to plain EI with
optimal (Rem.~\ref{Rem:optimalParameters}) hyperparameters in every
test case.

\section{CONCLUSION}

Both of our core contributions, length-scale cool down based on the
acquisition function, and the MGL kernel function, concern model
selection. From a higher-level perspective we proposed that in the
context of Bayesian Optimization we should select models differently
than in standard Machine Learning: Instead of selecting
hyperparameters based on maximum likelihood on the \emph{previous
  data} we should try to judge the implications of the choice of
hyperparameter on the \emph{future data}, e.g., via the acquisition
function. And instead of choosing a standard ``uninformed'' squared
exponential kernel we may want to choose a kernel function that
indirectly expresses our prior that model-based optimization is
efficient in local convex (non-isotropic) regions, while the global
length-scale characterizes where further local regions may are hidden.

We found that our novel concept of length-scale adpation outperforms
leave-one-out cross validation and even a-posteriori optimal hyperparameters
in the robust setting and has similar performance in the
nominal case.

Further, combining length-scale cool down and our novel MGL kernel
function with Expected Improvement shows in most benchmark problems
better performance while no assumptions on the model hyperparameters
were made beforehand. Furthermore we are not limited to EI which
enables the community to combine the length-scale adaption and model
with other acquisition functions, which potentially will lead to an
overall performance improvement regarding Bayesian optimization as we
believe.

\newpage

\subsubsection*{References}
{\def\section*#1{}
\bibliography{ArXiv_bib.bib}
}

\end{document}